\documentclass{article}

\usepackage[preprint,nonatbib]{neurips_data_2021}

\usepackage[utf8]{inputenc} % allow utf-8 input
\usepackage[T1]{fontenc}    % use 8-bit T1 fonts
\usepackage[colorlinks=true,citecolor=blue]{hyperref}       % hyperlinks
\usepackage{url}            % simple URL typesetting
\usepackage{booktabs}       % professional-quality tables
\usepackage{amsfonts}       % blackboard math symbols
\usepackage{nicefrac}       % compact symbols for 1/2, etc.
\usepackage{microtype}      % microtypography
\usepackage{xcolor}         % colors
\usepackage{times}
\usepackage{epsfig}
\usepackage{amsmath}
\usepackage{amssymb}
\usepackage{booktabs}
\usepackage{enumitem}
\usepackage{multirow}
\usepackage{hhline}
\usepackage[page]{appendix}
\usepackage[export]{adjustbox}
\usepackage{multirow}
\usepackage{float}
\restylefloat{table}
\newcommand{\im}{{ImageNet-21K-P}}
\usepackage[font=small,labelfont=bf]{caption}
\usepackage{wrapfig}
\usepackage{colortbl}
\usepackage{hhline}
\usepackage{vcell}

\DeclareFixedFont{\ttb}{T1}{txtt}{bx}{n}{8} % for bold
\DeclareFixedFont{\ttm}{T1}{txtt}{m}{n}{8}  % for normal
\usepackage{listings}
\usepackage{color}
\definecolor{deepblue}{rgb}{0,0,0.5}
\definecolor{deepred}{rgb}{0.6,0,0}
\definecolor{deepgreen}{rgb}{0,0.5,0}
\newcommand\pythonstyle{\lstset{
language=Python,
basicstyle=\ttm,
otherkeywords={self},             % Add keywords here
keywordstyle=\ttb\color{deepblue},
emph={split_logits_to_semantic_logits,calculate_KD_loss,estimate_teacher_confidence},          % Custom highlighting
emphstyle=\ttb\color{deepred},    % Custom highlighting style
stringstyle=\color{deepgreen},
frame=single,                         % Any extra options here
framexrightmargin=1pt,
showstringspaces=false            % 
}}
\lstnewenvironment{python_l}[1][]
{
\pythonstyle
\lstset{#1}
}
{}
\newcommand\pythoninline[1]{{\pythonstyle\lstinline!#1!}}
\usepackage{pythonhighlight}

\title{ImageNet-21K Pretraining for the Masses}

\author{%
Tal Ridnik \\
DAMO Academy, Alibaba Group\\
{\texttt{tal.ridnik@alibaba-inc.com }} \\
\And
Emanuel Ben-Baruch \\
DAMO Academy, Alibaba Group\\
{\texttt{emanuel.benbaruch@alibaba-inc.com }}
\And
Asaf Noy \\
DAMO Academy, Alibaba Group\\
{\texttt{asaf.noy@alibaba-inc.com }}
\And
Lihi Zelnik-Manor \\
DAMO Academy, Alibaba Group\\
{\texttt{lihi.zelnik@alibaba-inc.com }}
}

\begin{document}

\maketitle

\begin{abstract}
\label{sec:abstract}
ImageNet-1K serves as the primary dataset for pretraining deep learning models for computer vision tasks. ImageNet-21K dataset, which is bigger and more diverse, is used less frequently for pretraining, mainly due to its complexity, low accessibility, and underestimation of its added value.
This paper aims to close this gap, and make high-quality efficient pretraining on ImageNet-21K available for everyone.
Via a dedicated preprocessing stage, utilization of WordNet hierarchical structure, and a novel training scheme called semantic softmax, we show that various models significantly benefit from ImageNet-21K pretraining on numerous datasets and tasks, including small mobile-oriented models. 
We also show that we outperform previous ImageNet-21K pretraining schemes for prominent new models like ViT and Mixer.
Our proposed pretraining pipeline is efficient, accessible, and leads to SoTA reproducible results, from a publicly available dataset. The training code and pretrained models are available at: \url{ https://github.com/Alibaba-MIIL/ImageNet21K}
\end{abstract}

\section{Introduction}
ImageNet-1K dataset, introduced for the ILSVRC2012 visual recognition challenge \cite{ILSVRC15}, has been at the center of
modern advances in deep learning \cite{krizhevsky2012imagenet, he2016deep,sandler2018mobilenetv2}. ImageNet-1K serves as the main dataset for pretraining of models for computer-vision transfer learning \cite{tan2019efficientnet,lee2020compounding,howard2019searching}, and improving performances on ImageNet-1K is often seen as a litmus test for general applicability on 
downstream tasks \cite{kornblith2019better,zhang2017mixup,ridnik2021tresnet}.
ImageNet-1K is a subset of the full ImageNet dataset \cite{imagenet_cvpr09}, which consists of 14,197,122 images, divided into 21,841 classes. We shall refer to the full dataset as ImageNet-21K, following \cite{kolesnikovbig} (although other papers sometimes described it as ImageNet-22K \cite{codreanu2017scale}). ImageNet-1K was created by selecting a subset of 1.2M images from ImageNet-21K, that belong to 1000 mutually exclusive classes.
\begin{figure}[t]
\vspace{-0.3cm}
  \centering
  \includegraphics[scale=0.23,bb= 0 0 1676 504]{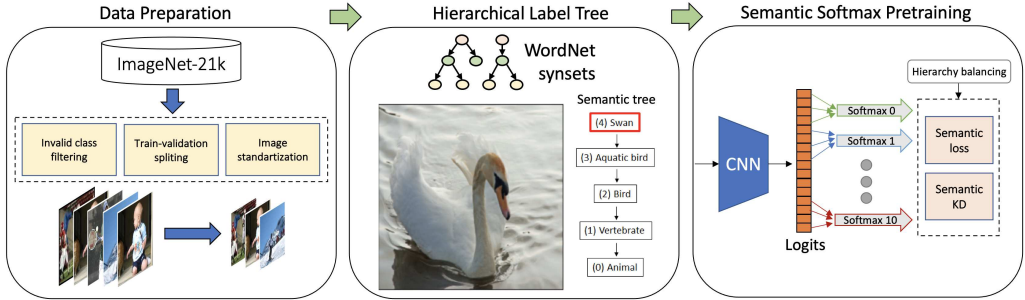}
  \caption{\textbf{Our end-to-end pretraining pipeline on ImageNet-21K.} We start with a dataset preparation and preprocessing stage. Via WordNet's synsets, we convert all the single-label inputs to semantic multi-labels, resulting in a semantic structure for ImageNet-21K, with $11$ possible hierarchies. For each hierarchy, we apply a dedicated softmax activation, and aggregate the losses with hierarchy balancing.}
  \label{fig:main_pic_3_stages}
\vspace{-0.3cm}
\end{figure} 

Even though some previous works showed that pretraining on ImageNet-21K could provide better downstream results for large models  \cite{kolesnikovbig,dosovitskiy2020image}, pretraining on ImageNet-1K remained far more popular. A main reason for this discrepancy is that ImageNet-21K labels are not mutually exclusive - the labels are taken from WordNet \cite{miller1995wordnet}, where each image is labeled with one label only, not necessarily at the highest possible hierarchy of WordNet semantic tree. For example, ImageNet-21K dataset contains the labels "chair" and "furniture". A picture, with an actual chair, can sometimes be labeled as "chair", but sometimes be labeled as the semantic parent of "chair", "furniture". This kind of tagging methodology complicates the training process, and makes evaluating models on ImageNet-21K less accurate. 
Other challenges of ImageNet-21K dataset are the lack of official train-validation split, the fact that training is longer than ImageNet-1K and requires highly efficient training schemes, and that the raw dataset is large - 1.3TB.
%

% While not as popular as ImageNet-1K, 
Several past works have used ImageNet-21K for pretraining, mostly in comparison to larger datasets, which are not publicly available, such as JFT-300M \cite{sun2017revisiting}.
\cite{mustafa2020deep} and \cite{puigcerver2020scalable} used ImageNet-21K and JFT-300M to train expert models according to the datasets hierarchies, and combined them to ensembles on downstream tasks; \cite{kolesnikovbig} and \cite{dosovitskiy2020image} compared pretraining JFT-300M to ImageNet-21K on large models such as ViT and ResNet-50x4. Many papers used these pretrained models for downstream tasks (e.g., \cite{zheng2020rethinking,neimark2021video,liu2021cptr, azizi2021big}). There are also works on ImageNet-21K that did not focus on pretraining: \cite{yun2021re} used extra (unlabled) data from ImageNet-21K to improve knowledge-distillation training on ImageNet-1K; \cite{dhillon2019baseline} used ImageNet-21k for testing few-shot learning; \cite{vijayanarasimhan2014deep} tested efficient softmax schemes on ImageNet-21k; \cite{goo2016taxonomy} tested pooling operations schemes on animal-oriented subset of ImageNet-21k.

However, previous works have not methodologically studied and optimized a pretraining process specifically for ImageNet-21K. Since this is a large-scale, high-quality, publicly available dataset, this kind of study can be highly beneficial to the community. We wish to close this gap in this work, and make efficient top-quality pretraining on ImageNet-21K accessible to all deep learning practitioners.

Our pretraining pipeline starts by preprocessing ImageNet-21K to ensure all classes have enough images for a meaningful learning, splitting the dataset to a standardized train-validation split, and resizing all images to reduce memory footprint. 
Using WordNet semantic tree \cite{miller1995wordnet}, we show that ImageNet-21K can be transformed into a (semantic) multi-label dataset. We thoroughly analyze the advantages and disadvantages of single-label and multi-label training. Extensive tests on downstream tasks show that multi-label pretraining does not improve results on downstream tasks, despite having more information per image. To effectively utilize the semantic data, we develop a novel training method, called \textit{semantic softmax}, which exploits the hierarchical structure of ImageNet-21K tagging to train the network over several semantic softmax layers, instead of the single layer. Using semantic softmax pretraining, we consistently outperform both single-label and multi-label pretraining on downstream tasks. By integrating semantic softmax into a dedicated semantic knowledge distillation loss, we further improved results.
The complete end-to-end pretraining pipeline appears in Figure \ref{fig:main_pic_3_stages}.

Using semantic softmax pretraining on ImageNet-21K we achieve significant improvement on numerous downstream tasks, compared to standard ImageNet-1K pretraining. Unlike previous works, which focused on pretraining of large models only \cite{kolesnikovbig}, we show that ImageNet-21K pretraining benefits a wide variety of models, from larger models like TResNet-L \cite{ridnik2021tresnet}, through medium-sized models like ResNet50 \cite{he2016deep}, and even small mobile-dedicated models like OFA-595 \cite{cai2019once} and MobileNetV3 \cite{howard2019searching}. Our proposed pretraining scheme also outperforms previous ImageNet-21K pretraining schemes that were used to trained MLP-based models like Vision-Transformer (ViT) \cite{dosovitskiy2020image} and Mixer \cite{tolstikhin2021mlp}. 
%
% Finally, we show that we obtain SoTA results on transfer learning to ImageNet-1K. For example, with ResNet50 model we achieves a top-1 accuracy of $82.0\%$.
%

The paper's contribution can be summarized as follows:
\begin{itemize}[leftmargin=0.4cm,topsep=-0.1cm]
  \setlength{\itemsep}{0.2pt}
  \setlength{\parskip}{0.2pt}
  \setlength{\parsep}{0.2pt}
  \item We develop a methodical preprocess procedure to transform raw ImageNet-21K into a viable dataset for efficient, high-quality pretraining. 
  \item Using WordNet semantic tree, we convert each (single) label to semantic multi labels, and compare the pretrain quality of two baseline methods: single-label and multi-label pretraining. We show that while a multi-label approach provides more information per image, it can have significant optimization drawbacks, resulting in inferior results on downstream tasks.
  \item We develop a novel training scheme called semantic softmax, which exploits the hierarchical structure of ImageNet-21K. With semantic softmax pretraining, we outperform both single-label and multi-label pretraining on downstream tasks. We further improve results by integrating semantic softmax into a dedicated semantic knowledge distillation scheme.
  \item Via extensive experimentations, we show that compared to ImageNet-1K pretraining, ImageNet-21K pretraining significantly improves downstream results for a wide variety of architectures, include mobile-oriented ones. 
  In addition, our ImageNet-21K pretraining scheme consistently outperforms previous ImageNet-21K pretraining schemes for prominent new models like ViT and Mixer.
\end{itemize}
% Our preprocessing scheme, training code and pretrained models are all publicly available.

\section{Dataset Preparation}
\subsection{Preprocessing ImageNet-21K}
\label{sec:pre_processing}
Our preprocessing stage  consists of three steps, as described in Figure \ref{fig:main_pic_3_stages} (leftmost image): (1) invalid classes cleaning, (2) creating a validation set, (3) image resizing. Details are as follows:
\\
\textbf{Step 1 - cleaning invalid classes:}
the original ImageNet-21K dataset \cite{imagenet_cvpr09} consists of 14,197,122 images, each tagged in a single-label fashion by one of 21,841 possible classes. The dataset has no official train-validation split, and the classes are not well-balanced - some classes contain only 1-10 samples, while others contain thousands of samples. Classes with few samples cannot be learned efficiently, and may hinder the entire training process and hurt the pretrain quality \cite{huh2016makes}. Hence we start our preprocessing stage by removing infrequent classes, with less than $500$ labels.
After this stage, the dataset contains 12,358,688 images from 11,221 classes. Notice that the cleaning process reduced the number of total classes by half, but removed only $13\%$ of the original pictures.
\\
\textbf{Step 2 - validation split:}
we allocate $50$ images per class for a standardized validation split, that can be used for future benchmarks and comparisons. 
\\
\textbf{Step 3 - image resizing:}
ImageNet-1K training usually uses \textit{crop-resizing} \cite{howard2020fastai} which favours loading the original images at full resolution and resizing them on-the-fly.
To make ImageNet-21K dataset more accessible and accelerate training, we resized during the preprocessing stage all the images to $224$ resolution (equivalent to\textit{ squish-resizing} \cite{howard2020fastai}). While somewhat limiting scale augmentations, this stage significantly reduces the dataset's memory footprint, from 1.3TB to 250GB, and makes loading the data during training faster.

After finishing the preprocessing stage, we kept only valid classes, produced a standardized train-validation split, and significantly reduced the dataset size.
We shall name this processed dataset \textbf{\im} (P for Processed).

\subsection{Utilizing Semantic Data}
\label{utilizing_semantic_data}
\begin{figure}[!t]
\vspace{-0.2cm}
    \begin{minipage}[t]{\textwidth}
        \begin{minipage}{0.48\textwidth}
            % \centering
            \includegraphics[scale=.4,bb=0 0 475 257]{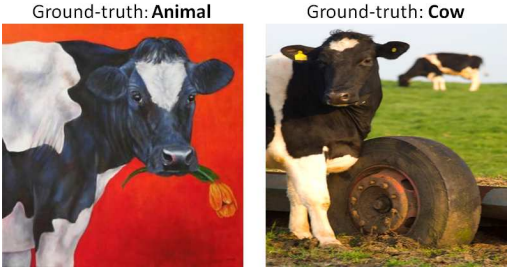}
            \captionof{figure}{\textbf{Example of inconsistent tagging in ImageNet-21K dataset.} Two pictures containing the same animal were labeled differently.}
            \label{fig:inconsistent_tagging}
        \end{minipage}
        \hfill
        \begin{minipage}{0.48\textwidth}
            % \centering
            \begin{table}[H]
\centering
\begin{tabular}{|c|c|} 
\hline
Hierarchy & Example Classes                                                                        \\ 
\hline
0         & \begin{tabular}[c]{@{}c@{}}person, animal, plant,\\~food, artifact\end{tabular}        \\ 
\hline
1         & \begin{tabular}[c]{@{}c@{}}domestic animal, \\basketball court, clothing\end{tabular}  \\ 
\hline
...       &                                                                                        \\ 
\hline
6         & \begin{tabular}[c]{@{}c@{}}whitetip shark, ortolan, \\grey kingbird\end{tabular}       \\
\hline
\end{tabular}
\caption{\textbf{Examples of classes from different \im{} hierarchies.}}
\label{table:example_classes}
\end{table}
        \end{minipage}
    \end{minipage}
\vspace{-0.3cm}
\end{figure}

We now wish to analyze the semantic structure of \im{} dataset. This structure will enable us to better understand \im{} tagging methodology, and employ and compare different pretraining schemes.
\vspace{-0.2cm}
\paragraph{From single labels to semantic multi labels}
Each image in the original ImageNet-21K dataset was labeled with a single label, that belongs to WordNet synset \cite{miller1995wordnet}. Using the  WordNet synset hyponym (subtype) and hypernym (supertype) relations, we can obtain for each class its parent class, if exists, and a list of child classes, if exists. When applying the parenthood relation recursively, we can build a semantic tree, that enables us to transform \im{} dataset into a multi-label dataset, where each image is associated with several labels - the original label, and also its parent class, parent-of-parent class, and so on. Example is given in Figure \ref{fig:main_pic_3_stages} (middle image) -
the original image was labeled as 'swan', but by utilizing the semantic tree, we can produce a list of semantic labels for the image - 'animal, vertebrate, bird, aquatic bird, swan'.
Notice that the labels are sorted by hierarchy: 'animal' label belongs  to hierarchy $0$, while 'swan' label belongs to hierarchy $4$. A label from hierarchy $k$ has $k$ ancestors.
\vspace{-0.2cm}
\paragraph{Understanding the inconsistent tagging methodology}
The semantic structure of ImageNet-21K enables us to understand its tagging methodology better. According to the stated tagging methodology of ImageNet-21K \cite{imagenet_cvpr09}, we are not guaranteed that each image was labeled at the highest possible hierarchy. An example is given in Figure \ref{fig:inconsistent_tagging}.
Two pictures, that contain the animal cow, were labeled differently - one with the label 'animal', the other with the label 'cow'. Notice that 'animal' is a semantic ancestor of 'cow' (cow $\rightarrow$ placental $\rightarrow$ mammal $\rightarrow$ vertebrate $\rightarrow$ animal).
This kind of incomplete tagging methodology, which is common in large datasets \cite{kuznetsova2020open,ngiam2018domain}, hinders and complicates the training process. 
A dedicated scheme that tackles this tagging methodology will be presented in section \ref{sec:semantic_softmax}.
\vspace{-0.2cm}
\paragraph{Semantic statistics}
\label{sec:semantic_statistics}
By using WordNet synsets, we can calculate for each class the number of ancestors it has - its hierarchy.
In total, our processed dataset, \im{}, has $11$ possible hierarchies. Example of classes from different hierarchies appears in Table \ref{table:example_classes}.
In Figure \ref{fig:class_hirarchy} in appendix \ref{appendix:classes_in_different_hierarchies} we present the number of classes per hierarchy.
We see that while there are $11$ possible hierarchies, the vast majority of classes belong to the lower hierarchies.

\section{Pretraining Schemes}
\label{sec:pretraining_schemes}\textit{}
In this section, we will review and analyze two baseline schemes for pretraining on ImageNet-21K-P: single-label and multi-label training. We will also develop a novel new scheme for pretraining on ImageNet-21K-P, \textit{semantic softmax}, and analyze its advantages over the baseline schemes.
\subsection{Single-label Training Scheme}
The straightforward way to pretrain on ImageNet-21K-P is to use the original (single) labels, apply softmax on the output logits, and use cross-entropy loss.
Our single-label training scheme is similar to common efficient training schemes on ImageNet-1K \cite{ridnik2021tresnet}, with minor adaptations to better handle the inconsistent tagging (Full training details appear in appendix \ref{appendix:single_label_training_details}). Since we aim for an efficient scheme with maximal throughput, we don't incorporate any tricks that might significantly increase training times. 
To further shorten the training times, we propose to initialize the models from standard ImageNet-1K training, and train on ImageNet-21K-P for $80$ epochs. On 8xV100 NVIDIA GPU machine,  mixed-precision training takes $40$ minutes per epoch for ResNet50 and TResNet-M architectures ($\sim$ $5000\frac{\textrm{img}}{\textrm{sec}}$), leading to a total training time of $54$ hours. Similar accuracies are obtained when doing random initialization, but training the models longer - $140$ epochs.
% Our single-label training scheme is similar to the common efficient training schemes on ImageNet-1K \cite{ridnik2021tresnet}, with three main adaptations:
% \begin{enumerate}[leftmargin=0.4cm]
% \setlength{\itemsep}{0.2pt}
% \setlength{\parskip}{0.2pt}
% \setlength{\parsep}{0.2pt}
% \item Due to the incomplete tagging of \im{}, we increase label-smooth factor from $0.1$ to $0.2$, to better handle ground-truth inconsistencies.
% \item As explained in section \ref{sec:pre_processing}, we use squish-resizing instead of crop-resizing.
% \item To shorten the training times, we initialize our model from ImageNet-1K training, and train on ImageNet-21K for $80$ epochs. On 8xV100 NVIDIA GPU machine, training with mixed-precision takes $40$ minutes per epoch for ResNet50 and TResNet-M architectures ($\sim$ $5000\frac{\textrm{img}}{\textrm{sec}}$), leading to a total training time of $54$ hours.
% \end{enumerate}
% Since we aim for an efficient scheme with maximal throughput, our training scheme does not incorporate any tricks that might significantly increase training times.
% Full training details appear in appendix \ref{appendix:single_label_training_details}.
%
\\
\\
\noindent
\textbf{Pros of using single-label training} 
\begin{itemize}[leftmargin=0.4cm]
  \setlength{\itemsep}{0.2pt}
  \setlength{\parskip}{0.2pt}
  \setlength{\parsep}{0.2pt}
\item \textbf{Well-balanced dataset} - with single-label training on \im{}, the dataset is well-balanced, meaning each class appears, roughly, the same number of times.
\item  \textbf{Single-loss training} - training with a softmax (a single loss) makes convergence easy and efficient, and avoids many optimization problems associated with multi-loss learning, such as different gradient magnitudes and gradient interference  \cite{yu2020gradient, chen2018gradnorm, crawshaw2020multi}. 
\end{itemize} 
\noindent
\textbf{Cons of using single-label training} 
\begin{itemize}[leftmargin=0.4cm]
  \setlength{\itemsep}{0.2pt}
  \setlength{\parskip}{0.2pt}
  \setlength{\parsep}{0.2pt}
\item \textbf{Inconsistent tagging} - due to the tagging methodology of \im{}, where we are not guaranteed that an image was labeled at the highest possible hierarchy, ground-truth labels are inherently inconsistent. Pictures, containing the same object, can appear with different single-label tagging (see Figure \ref{fig:inconsistent_tagging} for example).
\item \textbf{No semantic data} - during training, we are not presenting semantic data via the single-label ground-truth.
\end{itemize}

\subsection{Multi-label Training Scheme}
\label{multi_label_trainig_scheme}
Using the semantic tree, we can convert any (single) label to semantic multi labels, and train our models on \im{} in a multi-label fashion, expecting that the additional semantic information per image will improve the pretrain quality.
As commonly done in multi-label classification \cite{ben2020asymmetric}, we reduce the problem  to a series of binary classification tasks.
Given $N$ labels, the base network outputs one logit per label, $z_{n}$, and each logit is independently activated by a sigmoid function $\sigma(z_{n})$. Let's denote $y_{n}$ as the ground-truth for class $n$. The total classification loss, $L_{\text{tot}}$, is obtained by aggregating a binary loss from the $N$ labels: 
\begin{equation}\label{eq:loss}
        % L = \Sigma_{k=1}L_{k}(\sigma(z_{k}), y_{k})
        L_{\text{tot}} = \sum_{n=1}^{N}L\left(\sigma(z_{n}), y_{n}\right).
\end{equation}
Eq. \ref{eq:loss} formalizes multi-label classification as a multi-task problem. Since we have a large number of classes ($11,221$), this is an extreme multi-task case.
For training, we adopted the high-quality training scheme described in \cite{ben2020asymmetric}, that provided state-of-the-art results on large-scale multi-label datasets such as Open Images \cite{kuznetsova2020open}. Full training details appear in appendix \ref{appendix:multi_label_training_details}.
\\
\\
\noindent
\textbf{Pros of using multi-label training}
\begin{itemize}[leftmargin=0.4cm]
  \setlength{\itemsep}{0.8pt}
  \setlength{\parskip}{0.2pt}
  \setlength{\parsep}{0.2pt}
  \item \textbf{More information per image} - we present for each image all the available semantic labels.
  \item \textbf{Tagging and metrics are more accurate} - if an image was originally given a single label at hierarchy k, with multi-label training we are guaranteed that all ground-truth labels at hierarchies 0 to k are accurate.
  Hence, multi-label training partly mitigates the inconsistent tagging problem, and makes training metrics more accurate and reflective than single-label training. 
\end{itemize}

\noindent
\textbf{Cons of using multi-label training}
\begin{itemize}[leftmargin=0.4cm]
  \setlength{\itemsep}{0.8pt}
  \setlength{\parskip}{0.2pt}
  \setlength{\parsep}{0.2pt}
\item \textbf{Extreme multi-tasking} - with multi-label training, each class is learned separately (sigmoids instead of softmax). This extreme multi-task learning makes the optimization process harder and less efficient, and may cause convergences to a local minimum \cite{yu2020gradient, chen2018gradnorm, du2018adapting}.
\item \textbf{Extreme imbalancing} - as a multi-label dataset with many classes, \im{} suffers from a large positive-negative imbalance \cite{ben2020asymmetric}. In addition, due to the semantic structure, multi-label training is hindered by a large class imbalance \cite{johnson2019survey} - on average, classes from a lower hierarchy will appear far more frequent than classes from a higher hierarchy. 
\end{itemize}
In appendices \ref{appendix:upstream_results_multi} and \ref{appendix:multi_label_losses_comparison} we show that for multi-label training, ASL loss \cite{ben2020asymmetric}, that was designed to cope with large positive-negative imbalancing, significantly outperforms cross-entropy loss, both on upstream and downstream tasks. This supports our analysis of extreme imbalancing as a major optimization challenge of multi-label training. Notice that we also list extreme multi-tasking as another optimization pitfall of multi-label training, and a dedicated scheme for dealing with it might further improve results. However, most methods that tackle multi-task learning, such as GradNorm \cite{chen2018gradnorm} and PCGrad \cite{yu2020gradient}, require computation of gradients for each class separately. This is computationally infeasible for a dataset with a large number of classes, such as \im{}.

\subsection{Semantic Softmax Training Scheme}
\label{sec:semantic_softmax}
% \subsubsection{Motivation}
Our goal is to develop a dedicated training scheme that utilizes the advantages of both the single-label and the multi-label training. Specifically, our scheme should present for each input image all the available semantic labels, but use softmax activations instead of independent sigmoids to avoid extreme multi-tasking. 
We also want to have fully accurate ground-truth and training metrics, and provide the network direct data on the semantic hierarchies (this is not achieved even in multi-label training, the hierarchical structure there is implicit). In addition, the scheme should remain efficient in terms of training times.
% \begin{itemize}[leftmargin=0.4cm]
%   \setlength{\itemsep}{0.2pt}
%   \setlength{\parskip}{0.2pt}
%   \setlength{\parsep}{0.2pt}
%   \item No hiding of data from the network - present for each input image all the available semantic labels.
%   \item Use softmax activations instead of independent sigmoids, to avoid extreme multi-tasking and extreme positive-negative imbalancing
%   \item Provide the network direct data on the semantic hierarchies. Notice that this is not achieved even in multi-label training, the hierarchical structure is implicit there.
%   \item Have fully accurate ground-truth and training metrics.
%   \item Remain an efficient scheme in terms of training times.
% %   \item Improve downstream results.
% \end{itemize}

%
\begin{wrapfigure}{R}{0.5\textwidth}
  \centering
  \includegraphics[width=0.48\textwidth]{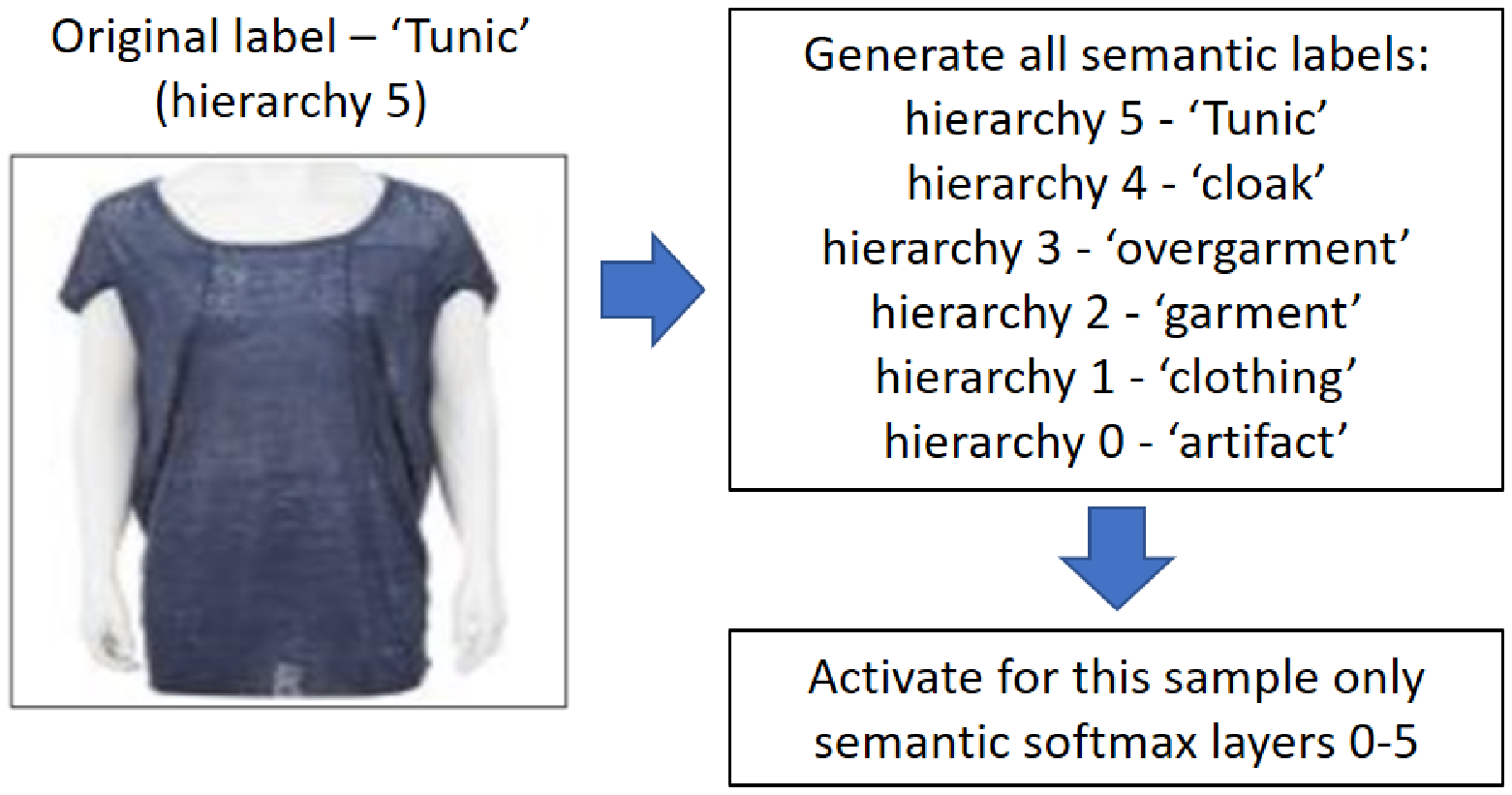}
  \caption{\textbf{Gradient propagation logic of semantic softmax training.}}
  \label{fig:relevant_layers_semantic_softmax_png}
  \vspace{-4mm}
\end{wrapfigure}
\vspace{-0.2cm}
\paragraph{Semantic softmax formulation}
To meet these goals, we develop a new training scheme called \textit{semantic softmax} training.
As we saw in section \ref{sec:semantic_statistics}, each label in \im{} can belong to one of $11$ possible hierarchies. By definition, for each hierarchy there can be only one ground-truth label per input image. Hence, instead of single-label training with a single softmax, we shall have $11$ softmax layers, for the $11$ different hierarchies. Each softmax will sample the relevant logits from the corresponding hierarchy, as shown in Figure \ref{fig:main_pic_3_stages} (rightmost image).
To deal with the partial tagging of \im{}, not all softmax layers will propagate gradients from each sample. Instead, we will activate only softmax layers from the relevant hierarchies. An example is given in Figure \ref{fig:relevant_layers_semantic_softmax_png} - the original image had a label from hierarchy $5$. We transform it to $6$ semantic ground-truth labels, for hierarchies 0-5, and activate only the $6$ first semantic softmax layers (only activated layers will propagate gradients).
Compared to single-label and multi-label schemes, semantic softmax training scheme has the following advantages:
\vspace{-0.2cm}
\begin{enumerate}[leftmargin=0.9cm]
  \setlength{\itemsep}{0.2pt}
  \setlength{\parskip}{0.2pt}
  \setlength{\parsep}{0.2pt}
  \item We avoid extreme multi-tasking ($11,221$ uncoupled losses in multi-label training). Instead, we have only $11$ losses, as the number of softmax layers.
  \item We present for each input image all the possible semantic labels. The loss scheme even provides direct data on the hierarchical structure.
  \item Unlike single-label and multi-label training, semantic softmax ground-truth and training metrics are fully accurate. If a sample has no labels at hierarchy k, we don't propagate gradients from the kth softmax during training, and ignore that hierarchy for metrics calculation (A dedicated metrics for semantic softmax training is defined in appendix \ref{appendix:semantic_upstream}).
  \item Calculating several softmax activations instead of a single one has negligible overhead, and in practice training times are similar to single-label training.
%   \item No approximation or unjustified assumptions.
\end{enumerate}
\vspace{-0.2cm}
\paragraph{Weighting the different softmax layers}
For each input image we have K losses (11). As commonly done in multi-task training \cite{chen2018gradnorm}, we need to aggregate them to a single loss. A naive solution will be to sum them:
$ L_{\text{tot}} = \sum_{k=0}^{K-1}L_{k}$
% \begin{equation}\label{eq:semantic_loss_naive}
%         L_{\text{tot}} = \sum_{k=0}^{K-1}L_{k}.
% \end{equation}
where $L_{k}$, the loss per softmax layer, is zero when the layer is not activated. 
However, this formulation ignores the fact that softmax layers at lower hierarchies will be activated much more frequently than softmax layers at higher hierarchies, resulting in over-emphasizing
of classes from lower hierarchies. To account for this imbalancing, we propose a balancing logic: let $N_{j}$ be the total number of classes in hierarchy j (as presented in Figure \ref{fig:class_hirarchy}). Due to the semantic structure, the relative number of occurrences of hierarchy k in the loss function will be: 
\begin{equation}\label{eq:semantic_loss_relative}
        O_{\text{k}} = \sum_{j=0}^{k-1}N_{j}
\end{equation}
Hence, to balance the contribution of different hierarchies we can use a normalization factor $W_{k}=\frac{1}{O_{k}}$, and obtain a balanced aggregation loss, that will be used for semantic softmax training:
\begin{equation}\label{eq:semantic_loss_balanced}
        L_{\text{tot}} = \sum_{k=0}^{K-1}W_{k}L_{k}
\end{equation}
\subsection{Semantic Knowledge Distillation}
Knowledge distillation (KD) is a known method to improve not only upstream, but also downstream results \cite{xie2020self, yan2020clusterfit, yun2021re}. We want to combine our semantic softmax scheme with KD training - \textit{semantic KD}. In addition  to the general benefit from KD training \cite{gou2021knowledge}, for ImageNet-21K-P semantic KD has an additional benefit - it can predict the missing tags that arise from the inconsistent tagging. For example, for the left picture in Figure \ref{fig:inconsistent_tagging}, the teacher model can predict the missing labels - 'cow, placental, mammal, vertebrate'. 
To implement semantic KD loss,
for each hierarchy we will calculate both the teacher and the student the corresponding probability distributions $\left\{T_{i}\right\}_{i=0}^{K-1}$, $\left\{S_{i}\right\}_{i=0}^{K-1}$.
The KD loss of hierarchy $i$ will be:
\begin{equation}
        L_{\text{KD}_{i}}=\text{KDLoss}(T_{i},S_{i})
\end{equation}
where KDLoss is a standard measurement for the distance between distributions, that can be chosen as Kullback-Leibler divergence \cite{gou2021knowledge, xie2020self}, or as MSE loss \cite{ba2013deep, tarvainen2017mean}. We have found that the latter converges faster, and used it.
A vanilla implementation for the total loss will be a simple sum of the losses from different hierarchies:
$
\label{eq:vanila_kd_loss}
        L_{\text{KD}}=\sum_{i=0}^{K-1}L_{\text{KD}{i}}
$.
% \begin{equation} \label{eq:vanila_kd_loss}
%         L_{\text{KD}}=\sum_{i=0}^{K-1}L_{\text{KD}{i}}.
% \end{equation}
However, this formulation assumes that all the hierarchies are relevant for each image. This is inaccurate - usually higher hierarchies represent subspecies of animals or plants, and are not applicable for a picture of a chair, for example. So we need to determine from the teacher predictions which hierarchies are relevant, and weigh the different losses accordingly. 
Let's assume that for each hierarchy we can calculate the teacher confidence level, $P_{i}$. A confidence-weighted KD loss will be:
\begin{equation} \label{eq:final_kd_loss}
        L_{\text{KD}}=\sum_{i=0}^{K-1}P_{i}L_{\text{KD}{i}}
\end{equation}
Eq. \ref{eq:final_kd_loss} is our proposed semantic KD loss. In appendix \ref{appendix:teacher_confidence_level} we present a method to calculate the teacher confidence level, $P_{i}$, from the teacher predictions, similar to \cite{xie2020self}.

\section{Experimental Study}
In this section, we will present upstream and downstream results for the different training schemes, and show that semantic softmax pretraining outperforms single-label and multi-label pretraining. We will also demonstrate how semantic KD further improves results on downstream tasks.
\subsection{Upstream Results}
In appendix \ref{appendix:upstream_results} we provide upstream results for the three training schemes.
Since each scheme has different training metrics, we cannot use these results to directly compare (pre)training quality.
%
% However, since we are using  \im{}, a standardized dataset with a fixed train-validation split, our upstream results can be used for future comparison and benchmarks (for each pretrain method).
%
\subsection{Downstream Results}
\label{sec:Downstream_Results}
To compare the pretrain quality of different training schemes, we will test our models via transfer learning.
To ensure that we are not overfitting a specific dataset or task, we chose a
wide variety of downstream datasets, from different computer-vision tasks. We also ensured that our downstream datasets represent a variety of domains, and have diverse sizes - from small datasets of thousands of images, to larger datasets with more than a million images.
For single-label classification, we transferred our models to ImageNet-1K \cite{krizhevsky2012imagenet}, iNaturalist 2019 \cite{van2018inaturalist}, CIFAR-100 \cite{krizhevsky2009learning} and Food 251 \cite{kaur2019foodx}. For multi-label classification, we transferred our models to MS-COCO \cite{lin2014microsoft} and Pascal-VOC \cite{everingham2010pascal} datasets. For video action recognition, we transferred our models to Kinetics 200 dataset \cite{kay2017kinetics}. In appendix \ref{appendix:downstream_datasets_training_details} we provide full training details on all downstream datasets.
\vspace{-0.2cm}
\paragraph{Comparing  different pretraining schemes}
In Table \ref{table:downstream_comparisons} we compare downstream results for three pretraining schemes: single-label, multi-label and semantic softmax.
\begin{table}[hbt!]
\centering
\arrayrulecolor{black}
\begin{tabular}{|c|c|c|c|c|} 
\hline
Dataset                           & \begin{tabular}[c]{@{}c@{}}Single\\Label\\Pretrain\end{tabular} & \begin{tabular}[c]{@{}c@{}}Mutli\\Label\\Pretrain\end{tabular} & \begin{tabular}[c]{@{}c@{}}Semantic\\Softmax\\Pretrain\end{tabular} & \begin{tabular}[c]{@{}c@{}}Semantic\\Softmax\\Pretrain + KD\end{tabular}  \\ 
\hhline{|=====|}
ImageNet1K\textsuperscript{(1)}   & {\cellcolor[rgb]{0.863,0.863,0.863}}81.1                        & 81.0                                                           & {\cellcolor[rgb]{0.706,0.706,0.706}}81.4                           & {\cellcolor[rgb]{0.588,0.588,0.588}}82.2                                 \\ 
\hline
iNaturalist\textsuperscript{(1)}  & {\cellcolor[rgb]{0.863,0.863,0.863}}71.5                        & 71.0                                                           & {\cellcolor[rgb]{0.706,0.706,0.706}}72.0                           & {\cellcolor[rgb]{0.588,0.588,0.588}}72.7                                 \\ 
\hline
Food 251\textsuperscript{(1) }    & {\cellcolor[rgb]{0.863,0.863,0.863}}75.4                        & 75.2                                                           & {\cellcolor[rgb]{0.706,0.706,0.706}}75.8                           & {\cellcolor[rgb]{0.588,0.588,0.588}}76.1                                 \\ 
\hline
CIFAR 100\textsuperscript{(1) }   & 89.5                                                            & {\cellcolor[rgb]{0.706,0.706,0.706}}90.6                       & {\cellcolor[rgb]{0.863,0.863,0.863}}90.4                           & {\cellcolor[rgb]{0.588,0.588,0.588}}91.7                                 \\ 
\hline
MS-COCO\textsuperscript{(2)}      & {\cellcolor[rgb]{0.863,0.863,0.863}}80.8                        & 80.6                                                           & {\cellcolor[rgb]{0.706,0.706,0.706}}81.3                           & {\cellcolor[rgb]{0.588,0.588,0.588}}82.2                                 \\ 
\hline
Pascal-VOC\textsuperscript{(2)}   & {\cellcolor[rgb]{0.863,0.863,0.863}}88.1                        & 87.9                                                           & {\cellcolor[rgb]{0.706,0.706,0.706}}89.7                           & {\cellcolor[rgb]{0.588,0.588,0.588}}89.8                                 \\ 
\hline
Kinetics 200\textsuperscript{(3)} & {\cellcolor[rgb]{0.863,0.863,0.863}}81.9                        & {\cellcolor[rgb]{0.863,0.863,0.863}}81.9                       & {\cellcolor[rgb]{0.706,0.706,0.706}}83.0                           & {\cellcolor[rgb]{0.588,0.588,0.588}}84.4                                 \\
\hline
\end{tabular}

\caption{\textbf{Comparing downstream results for different pretraining schemes.} Darker cell color means better score. Dataset types and metrics: (1) - single-label, top-1 Acc.[$\%$] ; (2) - multi-label, mAP [$\%$]; (3) - action recognition, top-1 Acc. [$\%$].}
\label{table:downstream_comparisons}

\end{table}

We see that on $6$ out of $7$ datasets tested, semantic softmax pretraining outperforms both single-label and multi-label pretraining. 
In addition, we see from Table \ref{table:downstream_comparisons} that single-label pretraining performs better than multi-label pretraining (scores are higher on $5$ out of $7$ datasets tested).

These results support our analysis of the pros and cons of the different pretraining schemes from Section \ref{sec:pretraining_schemes}: with multi-label training, we have more information per input image, but the optimization process is less efficient due to extreme multi-tasking and extreme imbalancing. All-in-all, multi-label training does not improve downstream results. Single-label training, despite its shortcomings from the partial tagging methodology and the minimal information per image, provides a better pretraining baseline.
Semantic softmax scheme, which utilizes semantic data without the optimization pitfalls of extreme multi-label training, outperforms both single-label and multi-label training.
\vspace{-0.2cm}
\paragraph{Semantic KD}
In Table \ref{table:downstream_comparisons} we also compare the downstream results of semantic softmax pretraining, with and without semantic KD.
We see that on all tasks and datasets tested, adding semantic KD to our pretraining process improves downstream results. Indeed the ability of semantic KD to fill in the missing tags and provide a smoother and more informative ground-truth is translated to better downstream results.
In appendix \ref{appendix:kd_comparisons_2} we compare single-label pretraining with KD, to semantic softmax pretraining with semantic KD, and show that the latter achieves better results on downstream tasks.

\section{Results}
In the previous chapters we developed a dedicated pretraining scheme for \im{} dataset, semantic softmax, and showed that it outperforms two baseline pretraining schemes, single-label and multi-label, in terms of downstream results. Now we wish to compare our semantic softmax pretraining on \im{} to other known pretraining schemes and pretraining datasets.

\subsection{Comparison to Other ImageNet-21K Pretraining Schemes}
We want to compare our proposed training scheme to other ImageNet-21K training schemes from the literature. However, to the best of our knowledge, no previous works have published their upstream results on ImageNet-21K, or shared thorough details about their training scheme or preprocessing stage. Recently, prominent new models called ViT \cite{dosovitskiy2020image} and Mixer \cite{tolstikhin2021mlp} were published, and official pretrained weights were released \cite{vit2021models}. In Table \ref{table:vit_comparison} we compare downstream results when using the official ImageNet-21K weights, and when using weights from semantic softmax pretraining.
\begin{table}[hbt!]
\centering
\setlength\tabcolsep{5.4pt} % default value: 6pt

\begin{tabular}{|c|c|c|c|c|c|} 
\cline{2-6}
\multicolumn{1}{l|}{\vcell{}}         & \multicolumn{2}{c|}{\vcell{ViT-B-16}}                                                                                                                              & \multicolumn{1}{l|}{\vcell{}}         & \multicolumn{2}{c|}{\vcell{Mixer-B-16}}                                                                                                                             \\[-\rowheight]
\multicolumn{1}{l|}{\printcellbottom} & \multicolumn{2}{c|}{\printcellbottom}                                                                                                                              & \multicolumn{1}{l|}{\printcellbottom} & \multicolumn{2}{c|}{\printcellbottom}                                                                                                                               \\ 
\cline{1-3}\cline{5-6}
Dataset                               & \begin{tabular}[c]{@{}c@{}}Official\\ImageNet-21K \\Pretrain\end{tabular} & \begin{tabular}[c]{@{}c@{}}Our\\ImageNet-21K \\Pretrain\end{tabular} &                                       & \begin{tabular}[c]{@{}c@{}}Official\\ImageNet-21K \\Pretrain\end{tabular} & \begin{tabular}[c]{@{}c@{}}Our\\ImageNet-21K \\Pretrain\end{tabular}  \\ 
\cline{1-3}\cline{5-6}
ImageNet1K\textsuperscript{(1)}       & 83.3                                                                                       & \textbf{83.9 }                                                        &                                       & 79.7                                                                                       & \textbf{82.0}                                                          \\ 
\cline{1-3}\cline{5-6}
iNaturalist\textsuperscript{(1)}      & 71.7                                                                                       & \textbf{73.1 }                                                        &                                       & 62.2                                                                                       & \textbf{66.6}                                                          \\ 
\cline{1-3}\cline{5-6}
Food 251\textsuperscript{(1) }        & 74.6                                                                                       & \textbf{76.0 }                                                        &                                       & 69.9                                                                                       & \textbf{74.5}                                                          \\ 
\cline{1-3}\cline{5-6}
CIFAR 100\textsuperscript{(1) }       & 92.7                                                                                       & \textbf{94.2 }                                                        &                                       & 85.5                                                                                       & \textbf{92.3}                                                          \\ 
\cline{1-3}\cline{5-6}
MS-COCO\textsuperscript{(2)}          & 81.1                                                                                       & \textbf{82.6 }                                                        &                                       & 74.1                                                                                       & \textbf{80.9}                                                          \\ 
\cline{1-3}\cline{5-6}
Pascal-VOC\textsuperscript{(2)}       & 78.7                                                                                       & \textbf{93.1 }                                                        &                                       & 63.1                                                                                       & \textbf{88.6}                                                          \\ 
\cline{1-3}\cline{5-6}
Kinetics 200\textsuperscript{(3)}     & 82.7                                                                                       & \textbf{84.1 }                                                        &                                       & 79.3                                                                                       & \textbf{82.1}                                                          \\
\hline
\end{tabular}
\caption{\textbf{Comparing downstream results for different pretraining schemes.} Dataset types and metrics: (1) - single-label, top-1 Acc.[$\%$] ; (2) - multi-label, mAP [$\%$]; (3) - action recognition, top-1 Acc. [$\%$]. }
\label{table:vit_comparison}
\end{table}
\vspace{-0.15cm}
\begin{table}[hbt!]
\vspace{-0.2cm}
\centering
\begin{tabular}{|c||c|c||c|c||c|c||c|c||c|c||} 
\hline
\multirow{2}{*}{Dataset}          & \multicolumn{2}{c||}{MobileNetV3} & \multicolumn{2}{c||}{OFA595} & \multicolumn{2}{c||}{ResNet50} & \multicolumn{2}{c||}{TResNet-M} & \multicolumn{2}{c||}{TResNet-L}  \\ 
\cline{2-11}
                                  & 1K   & 21K                        & 1K   & 21K                    & 1K   & 21K                     & 1K   & 21K                      & 1K   & 21K                       \\ 
\hhline{|=::==::==::==::==::==:|}
iNaturalist\textsuperscript{(1)}  & 62.4 & \textbf{65.0 }             & 69.0 & \textbf{71.5 }         & 66.8 & \textbf{71.4 }          & 70.1 & \textbf{72.7 }           & 72.4 & \textbf{74.8}             \\ 
\hline
CIFAR100\textsuperscript{(1)}     & 86.7 & \textbf{88.5 }             & 88.3 & \textbf{90.3 }         & 86.8 & \textbf{90.3 }          & 89.5 & \textbf{91.7 }           & 90.2 & \textbf{92.5}             \\ 
\hline
Food 251\textsuperscript{(1)}     & 70.1 & \textbf{70.3 }             & 72.9 & \textbf{73.5 }         & 72.2 & \textbf{74.0 }          & 75.1 & \textbf{76.1 }           & 76.3 & \textbf{77.0}             \\ 
\hhline{|=::==::==::==::==::==:|}
MS-COCO\textsuperscript{(2)}      & 73.0 & \textbf{74.9 }             & 74.9 & \textbf{77.7 }         & 76.7 & \textbf{80.5 }          & 79.5 & \textbf{82.2 }           & 81.1 & \textbf{83.7}             \\ 
\hline
Pascal-VOC\textsuperscript{(2)}   & 72.1 & \textbf{72.4 }             & 72.4 & \textbf{81.5 }         & 86.9 & \textbf{87.9 }          & 85.8 & \textbf{89.8 }           & 88.2 & \textbf{92.5}             \\ 
\hhline{|=::==::==::==::==::==:|}
Kinetics200\textsuperscript{(3)} & 72.2 & \textbf{74.3 }             & 73.2 & \textbf{78.1 }         & 78.2 & \textbf{81.3 }          & 80.5 & \textbf{84.3 }           & 82.1 & \textbf{84.6}             \\
\hline
\end{tabular}
\caption{\textbf{Comparing downstream results for ImageNet-1K standard pretraining, and our proposed \im{} pretraining scheme.} (1) - single-label dataset, top-1 Acc [$\%$] metric; (2) - multi-label dataset, mAP [$\%$] metric; (3) - action recognition dataset, top-1 Acc [$\%$] metric. }
\label{table:final_comparison}
\end{table}
\vspace{-0.15cm}
\\
We see from Table \ref{table:vit_comparison} that our pretraining scheme significantly outperforms the official pretrain, on all downstream tasks tested.
Previous works have observed that MLP-based models can be harder and less stable to use in transfer learning since they don't have inherent translation inductive bias \cite{chen2021empirical, mosbach2020stability, liu2020understanding}. When using the official weights, we also noticed this phenomenon on some datasets (Pascal-VOC, for example). Using semantic softmax pretraining, the transfer learning training was more stable and robust, and reached higher accuracy.
% As transformers-for-vision models are gaining popularity, the ability to train them to top results, using a publicly available and efficient scheme like ours can be highly beneficial.

\subsection{Comparison to ImageNet-1K Pretraining}

In Table \ref{table:final_comparison} we compare downstream results, for different models, when using ImageNet-1K pretraining (taken from \cite{rw2019timm}), and when using our \im{} pertraining.
We can see that our pretraining scheme significantly outperforms standard ImageNet-1K pretraining on all datasets, for all models tested. For example, on iNaturalist dataset we improve the average top-1 accuracy by $2.9\%$. 

Notice that some previous works stated that pretraining on a large dataset benefits only large models \cite{kolesnikovbig,sun2017revisiting}. MobileNetV3 backbone, for example, has only $4.2$M parameters,
while ViT-B model has $85.6$M parameters. Previous works assumed that a large number of parameters, like ViT has, is needed to properly utilize pretraining on large datasets. 
%
% This is the reason why ImageNet-1K was the primary source for pretraining of small models.
%
However, we show consistently and significantly that even small mobile-oriented models, like MobileNetV3 and OFA-595, can benefit from pretraining on a large (publicly available) dataset like \im{}.
Due to their fast inference times and reduced heating, mobile-oriented models are used frequently for deployment. Hence, improving their downstream results by using better pretrain weights can enhance real-world products, without increasing training complexity or inference times.

\subsection{ImageNet-1K SoTA Results}
In Table \ref{table:sota_imagenet_1k} in appendix \ref{appendix:imagenet_1k_sota_results} we bring downstream results on ImageNet-1K for different models, when using \im{} semantic softmax pretraining.
To achieve top results, similar to previous works \cite{lee2020compounding,yun2021re,xie2020self}, we added standard knowledge distillation loss into our ImageNet-1K training. 
To the best of our knowledge, for all the models in Table \ref{table:sota_imagenet_1k} we achieve a new SoTA record (for input resolution $224$). Unlike previous top works, which used private datasets \cite{sun2017revisiting}, we are using a publicly available dataset for pretraining.
Note that the gap from the original reported accuracies is significant. For example, MobileNetV3 reported accuracy was $75.2\%$ \cite{howard2019searching} - we achieved $78.0\%$; ResNet50 reported accuracy was $76.0\%$ \cite{he2016deep} - we achieved $82.0\%$.

\subsection{Additional Comparisons and Results}
In appendix \ref{additional Ablation} we bring additional comparisons: (1) Comparison to Open Images pretraining; (2) Downstream results comparison on additional non-classification computer-vision tasks; (3)  Impact of different number of training samples on upstream results.

\section{Conclusion}
In this paper, we presented an end-to-end scheme for high-quality efficient pretraining on ImageNet-21K dataset.
We start by standardizing the dataset preprocessing stage. Then we show how we can transform ImageNet-21K dataset into a multi-label one, using WordNet semantics. Via extensive tests on downstream tasks, we demonstrate how single-label training outperforms multi-label training, despite having less information per image. We then develop a new training scheme, called semantic softmax, which utilizes ImageNet-21K hierarchical structure to outperform both single-label and multi-label training. We also integrate the semantic softmax scheme into a dedicated knowledge distillation loss to further improve results.
On a variety of computer vision datasets and tasks, different architectures significantly and consistently benefit from our pretraining scheme, compared to ImageNet-1K pretraining and previous ImageNet-21K pretraining schemes.
% We also show that our pretraining scheme outperforms previous pretraining schemes for prominent new models like ViT and Mixer.
%
\vspace{-0.2cm}
\paragraph{Broader Impact}
In the past, pretraining on ImageNet-21K was out of scope for the common deep learning practitioner.
With our proposed pipeline, high-quality efficient pretraining on ImageNet-21K will be more accessible to the deep learning community, enabling researchers to design new architectures and pretrain them to top results, without the need for massive computing resources or large-scale private datasets. In addition, our findings that even small mobile-oriented models significantly benefit from large-scale pretraining can be used to enhance real-world products. Finally, our improved pretraining scheme on ImageNet-21K can support prominent MLP-based models that require large-scale pretraining, like ViT and Mixer.

% \section*{References}
{\small
\bibliographystyle{ieee_fullname.bst}
\bibliography{egbib.bib}
}

% \section{Checklist}
% \input{checklist.tex}

\clearpage

\appendix
\begin{appendices}

\section{Number of Classes in Different Hierarchies}
\label{appendix:classes_in_different_hierarchies}
\begin{figure}[hbt!]
  \centering
  \includegraphics[scale=.6]{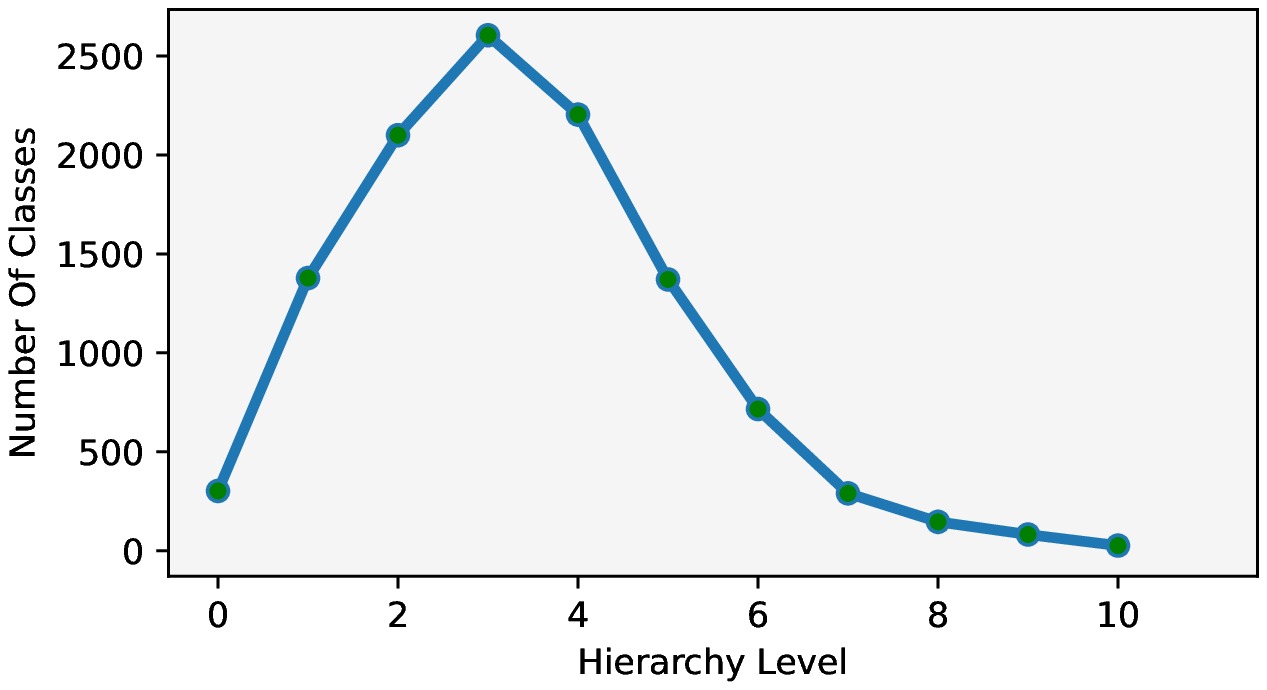}
  \caption{\textbf{Number of classes in different  hierarchies.}}
  \label{fig:class_hirarchy}
\end{figure} 

\section{Training Details}
\label{appendix:training_details}
\subsection{Single-label \im{} Training Details}
\label{appendix:single_label_training_details}
To better handle the ground-truth inconsistencies of \im{}, we increase the label-smooth factor from the common value of $0.1$ to $0.2$.
As explained in section \ref{sec:pre_processing}, we also use squish-resizing instead of crop-resizing.
We trained the models with input resolution $224$,  using an Adam optimizer with learning rate of 3e-4 and one-cycle policy \cite{smith2018disciplined}. When initializing our models from standard ImageNet-1K pretraining (pretraining weights taken from \cite{rw2019timm}), we found that $80$ epochs are enough for achieving strong pretrain results on \im{}.
For regularization, we used RandAugment \cite{cubuk2020randaugment}, Cutout \cite{Cutout}, Label-smoothing \cite{DBLP:journals/corr/SzegedyVISW15} and True-weight-decay \cite{loshchilov2017decoupled}. We observed that the common ImageNet statistics normalization \cite{lee2020compounding, tan2019efficientnet} does not improve the training accuracy, and instead normalized all the RGB channels to be between $0$ and $1$.
Unless stated otherwise, all runs and tests were done on TResNet-M architecture. 
On an 8xV100 NVIDIA GPU machine, training with mixed-precision takes $40$ minutes per epoch on ResNet50 and TResNet-M architectures ($\sim$ $5000\frac{\textrm{img}}{\textrm{sec}}$).

\subsection{Multi-label \im{} Training Details}
\label{appendix:multi_label_training_details}
For multi-label training, we convert each image single label input to semantic multi labels, as described in section \ref{utilizing_semantic_data}.
Multi-label training details are similar to single-label training (number of epochs, optimizer, augmentations, learning rate, models initialization and so on), and training times are also similar.
The main difference between single-label and multi-label training relies in the loss function: for multi-label training we tested $3$ loss functions, following \cite{ben2020asymmetric}: cross-entropy ($\gamma_{-}=\gamma_{+}=0$), focal loss ($\gamma_{-}=\gamma_{+}=2$) and ASL ($\gamma_{-}=4, \gamma_{+}=0$). For ASL, we tried different values of $\gamma_{-}$ to obtain the best mAP scores.

\section{Upstream Results}
\label{appendix:upstream_results}
As we have a standardized dataset with a fixed train-validation split, the training metrics for each pretraining method can be used for future benchmark and comparisons.

\subsection{Singe-label Upstream Results}
For single-label training, regular top-1 accuracy metric becomes somewhat irrelevant - if pictures with similar content have different ground-truth labels, the network has no clear "correct" answer. Top-5 accuracy metric is more representative, but still limited. 
Upstream results of single-label training are given in Table \ref{table:single_labels_scores}.
We can see that the top-1 accuracies obtained on \im{}, $37\%-46\%$, are significantly lower than the ones obtained on ImageNet-1K, $75\%-85\%$. This accuracy drop is mainly due to the semantic structure and inconsistent tagging methodology of \im{}.
However, as we take bigger and better architectures, we see from Table \ref{table:single_labels_scores} that the accuracies continue to improve, so we are not completely hindered by the inconsistent tagging.
\begin{table}[hbt!]
\centering
\begin{tabular}{|c|c|c|}
\hline
Model Name  & Top-1 Acc.  [\%] & Top-5 Acc.  [\%]  \\ 
\hline\hline
MobileNetV3 & 37.8           & 66.3             \\
\hline
ResNet50    & 42.2           & 72.0             \\
\hline
TResNet-M   & 45.3           & 75.2            \\
\hline
TResNet-L   & 45.5           & 75.6   \\ 
\hline
\end{tabular}
\caption{\textbf{Accuracy of different models in single-label training.}}
\label{table:single_labels_scores}
\end{table}

\subsection{Multi-label Upstream Results}
\label{appendix:upstream_results_multi}
For multi-label training, we will use the common micro and macro mAP accuracy \cite{ben2020asymmetric} as training metrics. However, due to the missing labels in the validation (and train) set, this metric also is not fully accurate.
In Table \ref{table:multi_label_losses_comparison} we compare the results for three possible loss functions for multi-label classification - cross-entropy, focal loss and ASL.
\begin{table}[hbt!]
\centering
\begin{tabular}{|c|c|c|}
\hline
Loss Type  & Micro-mAP [\%] & Macro-mAP  [\%]  \\ 
\hline\hline
Cross-Entropy & 47.3            & 73.9             \\
\hline
Focal loss    & 47.4            & 74.1             \\
\hline
ASL           & 48.5           & 74.7   \\
\hline
\end{tabular}
\caption{\textbf{Comparing different loss functions for multi-label classification on \im{}.}}
\label{table:multi_label_scores_imagenet_21k}
\end{table}
We see that ASL loss \cite{ben2020asymmetric}, that was designed to cope with large positive-negative imbalancing, outperform cross-entropy and focal loss. This is in agreement with our analysis in section \ref{multi_label_trainig_scheme}, where we identify extreme imbalancing as one of the optimization challenges that stems from multi-label training. 

\subsection{Semantic Softmax Upstream Results}
\label{appendix:semantic_upstream}
With semantic softmax training, we can calculate for each hierarchy its top-1 accuracy metric.
We can also calculate the total accuracy by weighting the different accuracies by the number of classes in each hierarchy (see Figure \ref{fig:class_hirarchy}).
Notice that we are not using classes above the maximal hierarchy for our metrics calculation. Hence, and unlike single-label and multi-label training, with semantic softmax our training metrics are fully accurate.

\begin{figure}[hbt!]
  \centering
  \includegraphics[scale=.55,bb=-0 -0 440 212]{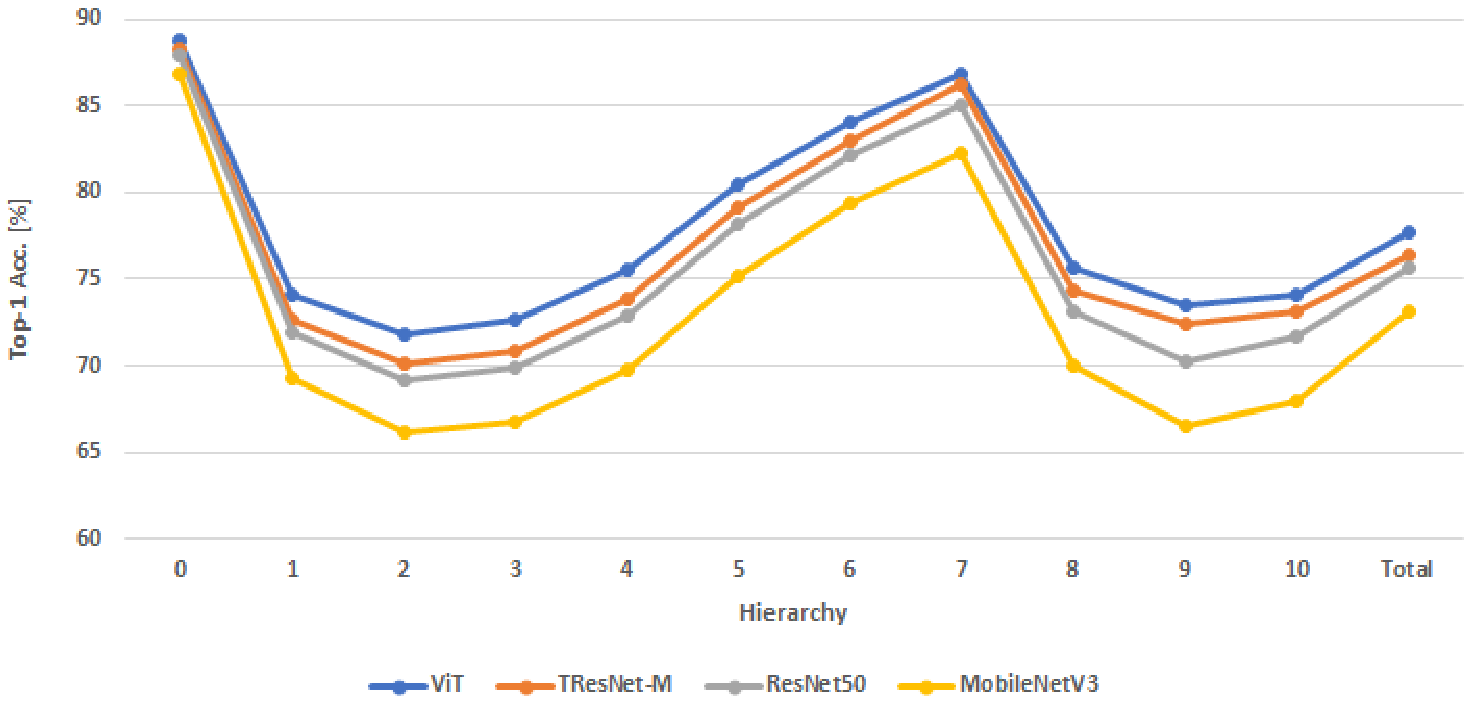}
  \caption{\textbf{Top-1 accuracies on different hierarchies.}}
  \label{fig:semantic_accuracies}
\end{figure}

In Figure \ref{fig:semantic_accuracies} we present the top-1 accuracies achieved by different models on different hierarchy levels, when trained with semantic softmax (with KD). 

\section{Downstream Datasets Training Details}
\label{appendix:downstream_datasets_training_details}
For single-label classification, our downstream datasets were
ImageNet-1K \cite{krizhevsky2012imagenet}, iNaturalist 2019 \cite{van2018inaturalist}, CIFAR-100 \cite{krizhevsky2009learning} and Food-251 \cite{kaur2019foodx}. For multi-label classification, our downstream datasets were MS-COCO \cite{lin2014microsoft} and Pascal-VOC \cite{everingham2010pascal}. For video action recognition, our downstream dataset was Kinetics-200 \cite{kay2017kinetics}.
\\
\paragraph{General details:}
\begin{itemize}[leftmargin=0.4cm]
  \setlength{\itemsep}{0.2pt}
  \setlength{\parskip}{0.2pt}
  \setlength{\parsep}{0.2pt}
  \item To minimize statistical uncertainty, for datasets with less than $150,000$ images (CIFAR-100, Food-251, MS-COCO, Pascal-VOC), we report result of averaging $3$ runs with different seeds.
  \item All results are reported for input resolution $224$.
  \item For all downstream datasets we used cutout of $0.5$, rand-Augment and true-weight-decay of 1e-4.
  \item All single-label datasets are trained with label-smooth of $0.1$
  \item Unless stated otherwise, dataset was trained for $40$ epochs with Adam optimizer, learning rate of 3e-4, one-cycle policy and and squish-resizing.
  
\end{itemize}
\paragraph{Specific dataset details:}
\begin{itemize}[leftmargin=0.4cm]
  \setlength{\itemsep}{0.2pt}
  \setlength{\parskip}{0.2pt}
  \setlength{\parsep}{0.2pt}
  \item ImageNet-1K - Since the dataset is bigger than the others, we finetuned our networks for $100$ epochs using SGD optimizer, and learning rate of 4e-4. We used crop-resizing with the common minimal crop factor of $0.08$.
  \item MS-COCO - We used ASL loss with $\gamma_{-}=4$.
  \item Pascal-VOC - We used ASL loss with $\gamma_{-}=4$, and learning rate of 5e-5.
  \item Kinetics-200 - we trained for $30$ epochs with learning rate of 8e-5. We used the training method described in \cite{sharir2021image}, with simple averaging of the embedding from each sample along the video.
\end{itemize}

\section{Downstream Results for Different Multi-label Losses}
\label{appendix:multi_label_losses_comparison}
In Table \ref{table:multi_label_losses_comparison} we compare downstream results when using multi-label pretraining with vanilla cross-entropy (CE) loss and ASL loss. We see that on all downstream datasets, pretraining with ASL leads to significantly better results
\begin{table}[hbt!]
\centering
\begin{tabular}{|c|c|c|} 
\hline
Dataset                           & \begin{tabular}[c]{@{}c@{}}Multi\\Label\\Pretrain\\(CE)\end{tabular} & \begin{tabular}[c]{@{}c@{}}Multi\\Label\\Pretrain \\(ASL)\end{tabular}  \\ 
\hline\hline
ImageNet1K\textsuperscript{(1)}   & 79.6                                                                 & \textbf{81.0 }                                                          \\ 
\hline
iNaturalist\textsuperscript{(1)}  & 69.4                                                                 & \textbf{71.0 }                                                          \\ 
\hline
Food 251\textsuperscript{(1)}     & 74.3                                                                 & \textbf{75.2 }                                                          \\ 
\hline
CIFAR 100\textsuperscript{(1)}    & 89.9                                                                 & \textbf{90.6}                                                           \\ 
\hline\hline
MS-COCO\textsuperscript{(2)}      & 79.1                                                                 & \textbf{80.6 }                                                          \\ 
\hline
Pascal-VOC\textsuperscript{(2)}   & 87.6                                                                 & \textbf{87.9 }                                                          \\ 
\hline\hline
Kinetics 200\textsuperscript{(3)} & 81.1                                                                 & \textbf{81.9}                                                           \\
\hline
\end{tabular}
\caption{\textbf{Comparing downstream results for different losses of multi-label pretraining.} Dataset types and metrics: (1) - single-label, top-1 Acc.[$\%$] ; (2) - multi-label, mAP [$\%$]; (3) - action recognition, top-1 Acc. [$\%$]. }
\label{table:multi_label_losses_comparison}
\end{table}

\section{Calculating Teacher Confidence}
\label{appendix:teacher_confidence_level}
Using the teacher prediction for hierarchy i and the semantic ground-truth, we want to  evaluate the teacher confidence level, $P_{i}$, so we can weight properly the contribution of different hierarchies in the KD loss.
Our proposed logic for calculating the teacher's (semantic) confidence is simple: 
\\
- If the ground-truth highest hierarchy is higher than i, set $P_{i}$ to 1. 
\\
- Else, calculate the sum probabilities of the top $5\%$ classes in the teacher prediction (we deliberately don't take only the probability of the highest class, to account for class similarities). 

\begin{figure}[hbt!]
  \centering
  \includegraphics[scale=.61,bb=-0 -0 396 180]{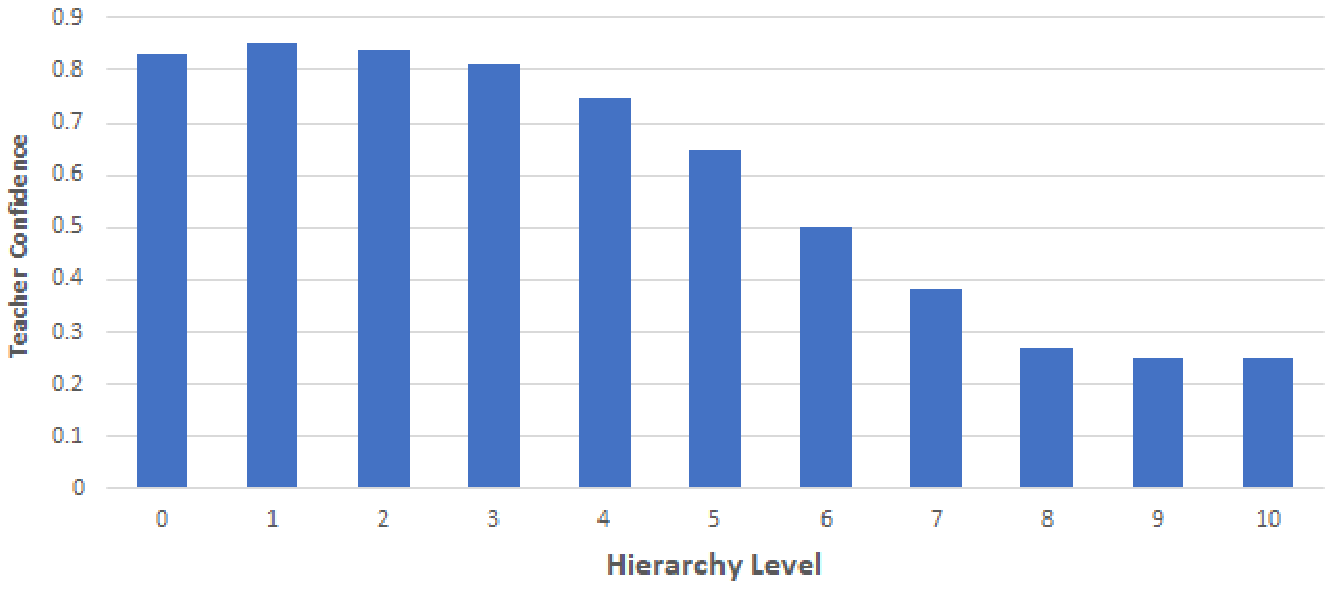}
  \caption{\textbf{Teacher average confidence levels for different hierarchies.}}
  \label{fig:teacher_confidene_png}
\end{figure}

In Figure \ref{fig:teacher_confidene_png} we present the teacher confidence level for different hierarchies, averaged over an epoch.
We can see that lower hierarchies have, in average, higher confidence levels. This stems from the fact that not all hierarchies are relevant for each image. For the picture in Figure \ref{fig:relevant_layers_semantic_softmax_png}, for example, only hierarchies 0-5 are relevant, so we expect the teacher will have low confidence for hierarchies higher than 5.

\section{Semantic KD Vs Regular KD}
\label{appendix:kd_comparisons_2}
\begin{table}[hbt!]
\centering
\refstepcounter{table}
\begin{tabular}{|c|c|c|} 
\hline
Dataset                           & \begin{tabular}[c]{@{}c@{}}Single\\Label \\+ KD\\Pretrain\end{tabular} & \begin{tabular}[c]{@{}c@{}}Sematic\\Softmax\\+ Semanic KD\\Pretrain\\\end{tabular}  \\ 
\hline\hline
ImageNet1K\textsuperscript{(1)}   & 81.5                                                                   & \textbf{82.2}                                                                       \\ 
\hline
iNaturalist\textsuperscript{(1)}  & 72.4                                                                   & \textbf{72.7}                                                                       \\ 
\hline
Food 251\textsuperscript{(1) }    & 76.0                                                                   & \textbf{76.1}                                                                       \\ 
\hline
CIFAR 100\textsuperscript{(1) }   & 91.0                                                                   & \textbf{91.7}                                                                       \\ 
\hline\hline
MS-COCO\textsuperscript{(2)}      & 81.6                                                                   & \textbf{82.2}                                                                       \\ 
\hline
Pascal-VOC\textsuperscript{(2)}   & 89.0                                                                   & \textbf{89.8}                                                                       \\ 
\hline\hline
Kinetics 200\textsuperscript{(3)} & 83.6                                                                   & \textbf{84.4}                                                                       \\
\hline
\end{tabular}
\caption{\textbf{Comparing KD with different schemes.} Dataset types and metrics: (1) - single-label, top-1 Acc.[$\%$] ; (2) - multi-label, mAP [$\%$]; (3) - action recognition, top-1 Acc. [$\%$]. }
\label{table:kd_comparisons_2}
\end{table}

\section{ImageNet-1K Transfer Learning Results}
\label{appendix:imagenet_1k_sota_results}
\begin{table}[hbt!]
\centering
\begin{tabular}{|c|c|} 
\hline
Model Name       & \begin{tabular}[c]{@{}c@{}}ImageNet-1K + KD \\Top-1 Acc. [\%]\end{tabular}  \\ 
\hline\hline
MobileNetV3 & 78.0                                                                        \\ 
\hline
OFA-595      & 81.0                                                                        \\ 
\hline
ResNet50     & 82.0                                                                        \\ 
\hline
Mixer-B-16     & 82.2                                                                \\ 
\hline
TResNet-M    & 83.1                                                                        \\ 
\hline
TResNet-L    & 83.9                                                                        \\ 
\hline
ViT-B-16     & 84.4                                                                        \\
\hline
\end{tabular}
\caption{\textbf{Transfer learning results On ImageNet-1K, when using \im{} pretraining.}}
\label{table:sota_imagenet_1k}
\end{table}

\section{ImageNet-21K-P - Winter21 Split}
For a fair comparison to previous works, the results in the article are based on the original ImageNet-21K images, i.e. we are using Fall11 release of ImageNet-21K (\textit{fall11-whole.tar} file), which contains all the original images and classes of ImageNet-21K. After we processed this release to create ImageNet-21K-P, we are left with a dataset that contains 11221 classes, where the train set has 11797632 samples and the test set has 561052 samples. We shall name this variant \textit{Fall11 ImageNet-21K-P}.

Recently, the official ImageNet site\footnote{www.image-net.org} used our pre-processing methodology to offer direct downloading of ImageNet-21K-P, based on a new release of ImageNet-21K - Winter21 (\textit{winter21-whole.tar} file). Compared to the original dataset, the Winter21 release removed some classes and samples. The Winter21 variant of ImageNet-21K-P is a dataset that contains 10450 classes, where the train set has 11060223  samples and the test set has 522500 samples. We shall name this variant \textit{Winter21 ImageNet-21K-P}.

For enabling future comparison and benchmarking, we report the upstream accuracies also on this new variant of ImageNet-21K-P:

\begin{table}[hbt!]
\centering
\begin{tabular}{|c|c|c|c|} 
\hline
                                                                  & \begin{tabular}[c]{@{}c@{}}Single-Label \\Training Acc. [\%]\end{tabular} & \begin{tabular}[c]{@{}c@{}}Multi-Label Training \\Macro-mAP [\%]\end{tabular} & \begin{tabular}[c]{@{}c@{}}Semantic Softmax \\Training Acc. [\%]\end{tabular}  \\ 
\hhline{|====|}
\begin{tabular}[c]{@{}c@{}}Fall11 \\ImageNet-21K-P\end{tabular}   & 45.3                                                                      & 74.7                                                                         & 75.6                                                                                           \\ 
\hline
\begin{tabular}[c]{@{}c@{}}Winter21 \\ImageNet-21K-P\end{tabular} & 47.3                                                                      & 78.7                                                                               & 77.7                                                                                           \\
\hline
\end{tabular}
\caption{\textbf{Upstream results, with different pretraining methods, for different variants of ImageNet-21K-P.} Tested model - TResNet-M.}
\end{table}

Note that the Winter21 variant of ImageNet-21K-P contains 10\% fewer classes and 6\% fewer images. In Table \ref{table:different_releases} we compare downstream results when using Winter21 and Fall11 variants of ImageNet-21K-P

\begin{table}[hbt!]
\centering
\begin{tabular}{|c|c|c|} 
\hline
Dataset                           & Fall11 ImageNet-21K-P & Winter21 ImageNet-21K-P  \\ 
\hhline{|===|}
ImageNet1K\textsuperscript{(1)}   & \textbf{81.4}         & 81.2                     \\ 
\hline
iNaturalist\textsuperscript{(1)}  & \textbf{72.0}         & 71.8                     \\ 
\hline
Food 251\textsuperscript{(1)}     & \textbf{75.8}         & 75.5                     \\ 
\hline
CIFAR 100\textsuperscript{(1)}    & 90.4                  & \textbf{90.5}            \\ 
\hhline{|===|}
MS-COCO\textsuperscript{(2)}      & \textbf{81.3}         & 81.1                     \\ 
\hline
Pascal-VOC\textsuperscript{(2)}   & 89.7                  & \textbf{90.1}            \\ 
\hhline{|===|}
Kinetics 200\textsuperscript{(3)} & \textbf{83.0}         & 82.8                     \\
\hline
\end{tabular}
\caption{\textbf{Comparing downstream results when using different variant of ImageNet-21K-P.} All results are for MTResNet model, with semantic softmax pretraining. Dataset types and metrics: (1) - single-label, top-1 Acc.[$\%$] ; (2) - multi-label, mAP [$\%$]; (3) - action recognition, top-1 Acc. [$\%$]. }
\label{table:different_releases}
\end{table}

We can see that compared to Fall11 variant, using Winter21 variant leads to a minor reduction in performances on downstream tasks.

\section{Additional Ablation Tests }
\label{additional Ablation}
In this section we will bring additional ablation tests and comparisons.
\subsection{Comparison to Pretraining on Open Images Dataset}
Open Images (v6) \cite{kuznetsova2018open} is a large scale multi-label dataset, which consists of $9$ million training images and $9600$ labels.
In Table \ref{table:open_images} we compare downstream results when using two different datasets for pretraining: ImageNet-21K (semantic softmax training) and Open Images (multi-label training).
\begin{table}[htbp]
\centering
\begin{tabular}{|c|c|c|} 
\hline
Dataset                           & \begin{tabular}[c]{@{}c@{}}ImageNet-21K \\Pretrain\end{tabular} & \begin{tabular}[c]{@{}c@{}}Open Images\\Pretrain\end{tabular}  \\ 
\hline
ImageNet1K\textsuperscript{(1)}   & \textbf{81.4 }                                                  & 81.0                                                           \\ 
\hline
iNaturalist\textsuperscript{(1)}  & \textbf{72.0 }                                                  & 70.7                                                            \\ 
\hline
Food 251\textsuperscript{(1) }    & \textbf{75.8 }                                                  & 74.8                                                           \\ 
\hline
CIFAR 100\textsuperscript{(1) }   & \textbf{90.4 }                                                  & 89.4                                                           \\ 
\hline
MS-COCO\textsuperscript{(2)}      & \textbf{81.3 }                                                  & 80.5                                                           \\ 
\hline
Pascal-VOC\textsuperscript{(2)}   & \textbf{89.7}                                                   & 89.6                                                           \\ 
\hline
Kinetics 200\textsuperscript{(3)} & \textbf{83.0}                                                   & 81.6                                                            \\
\hline
\end{tabular}
\caption{\textbf{Comparing ImageNet-21K pretraining to Open Images pretraining.} Downstream dataset types and metrics: (1) - single-label, top-1 Acc.[$\%$] ; (2) - multi-label, mAP [$\%$]; (3) - action recognition, top-1 Acc. [$\%$]. }
\label{table:open_images}
\end{table}

As we can see, ImageNet-21K pretraining consistently provides better downstream results than Open Images. A possible reason is that Open Images, as a multi-label dataset with large number of classes, suffers from the same multi-label optimization pitfalls we described in section \ref{multi_label_trainig_scheme}.

\subsection{Comparison on Additional Non-Classification Computer-Vision Tasks}
In Table \ref{table:ms_coco} and Table \ref{table:inria} we compare 1K and 21K pretraining on two additional computer-vision tasks: object detection (MS-COCO dataset) and image retrieval (INRIA holidays dataset).
\begin{table}[hbt!]
\centering
\begin{tabular}{|l|c|c|} 
\hline
          & 1K Pretraining & 21K Pretraining  \\ 
\hline
mAP [\%] & 42.9           & 44.3             \\
\hline
\end{tabular}
\caption{\textbf{Comparing downstream results on MS-COCO object detection dataset.} }
\label{table:ms_coco}
\end{table}
\begin{table}[hbt!]
\centering
\begin{tabular}{|l|c|c|} 
\hline
          & 1K Pretraining & 21K Pretraining  \\ 
\hline
mAP [\%] & 81.1           & 82.1             \\
\hline
\end{tabular}
\caption{\textbf{Comparing downstream results on on INRIA Holidays image retrieval dataset.} }
\label{table:inria}
\end{table}

We can see that also on non-classification tasks such as object detection and image retrieval, pretraining on ImageNet-21K translates to better downstream results than ImageNet-1K pretraining.

\subsection{Impact of Different Number of Training Samples}
\begin{figure}[hbt!]
  \centering
  \includegraphics[scale=.6]{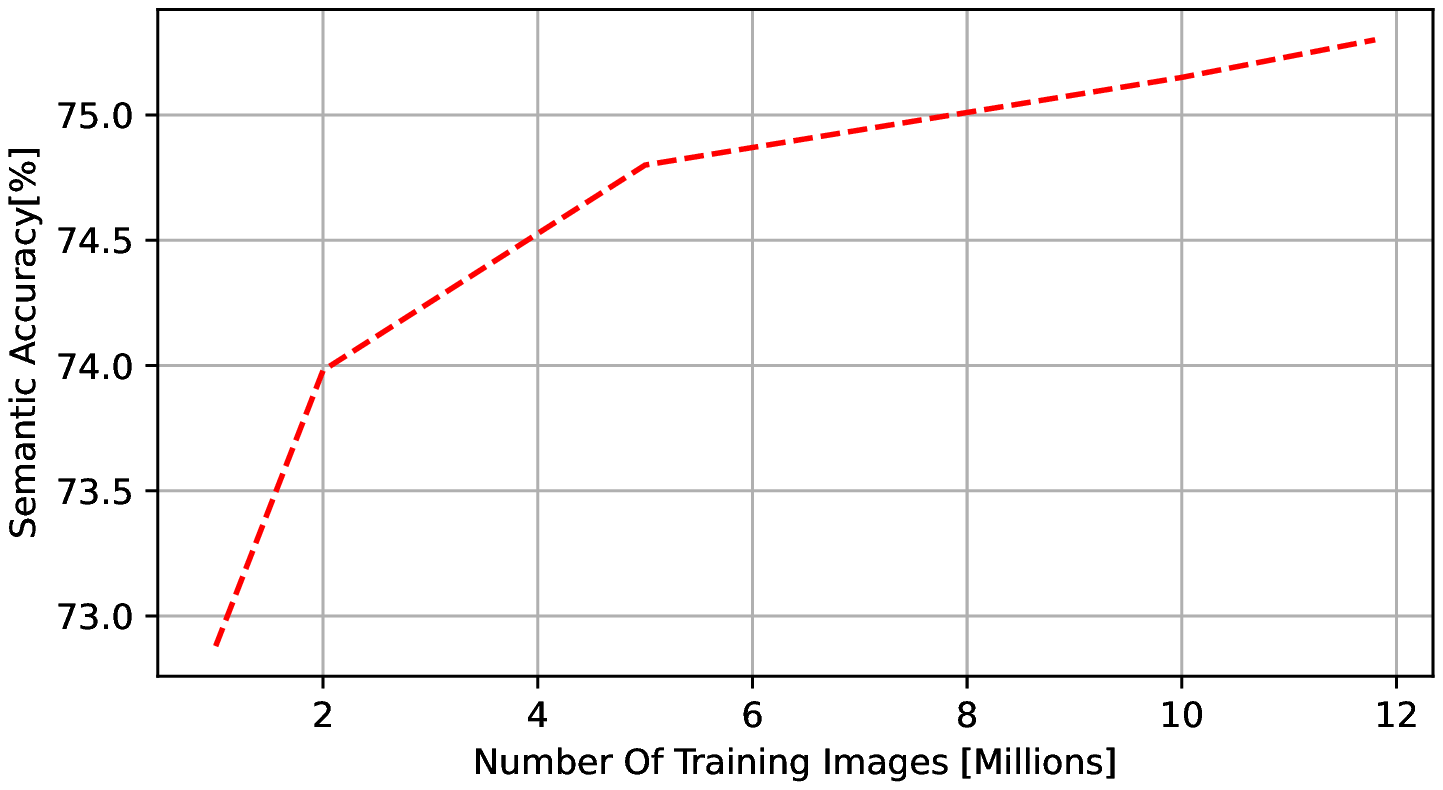}
  \caption{\textbf{Upstream results for different number of training images.}}
  \label{impact_number_samples}
\end{figure}

In Figure \ref{impact_number_samples} we test the impact of the number of training samples on on the upstream accuracies. As we can see, there is no saturation - more training images lead to better semantic accuracies.

\section{Pseudo-code}
In the following sections we will bring pseudo-code (PyTorch-style) to some components in our semantic softmax training scheme: logits sampling, KD calculation and estimating teacher confidence. 
\\
\\
\\
\subsection{Logits Sampling}
\begin{python_l}
def split_logits_to_semantic_logits(logits, hierarchy_indices_list):
    semantic_logit_list = []
    for i, ind in enumerate(hierarchy_indices_list):
        logits_i = logits[:, ind]
        semantic_logit_list.append(logits_i)
    return semantic_logit_list
\end{python_l}

\subsection{KD Logic}
\begin{python_l}
def calculate_KD_loss(input_student, input_teacher, hierarchy_indices_list):
    semantic_input_student = split_logits_to_semantic_logits(
    input_student, hierarchy_indices_list)
    semantic_input_teacher = split_logits_to_semantic_logits(
    input_teacher, hierarchy_indices_list)
    number_of_hierarchies = len(semantic_input_student)

    losses_list = []
    # scanning hirarchy_level_list
    for i in range(number_of_hierarchies):
        # converting to semantic logits
        inputs_student_i = semantic_input_student[i]
        inputs_teacher_i = semantic_input_teacher[i]

        # generating probs
        preds_student_i = stable_softmax(inputs_student_i)
        preds_teacher_i = stable_softmax(inputs_teacher_i)

        # weight MSE-KD distances according to teacher confidence
        loss_non_reduced = torch.nn.MSELoss(reduction='none')(preds_student_i,
        preds_teacher_i)
        weights_batch = estimate_teacher_confidence(preds_teacher_i)
        loss_weighted = loss_non_reduced * weights_batch.unsqueeze(1)
        losses_list.append(torch.sum(loss_weighted))

    return sum(losses_list)
\end{python_l}

\subsection{Teacher Confidence}
\begin{python_l}
def estimate_teacher_confidence(preds_teacher)
    with torch.no_grad():
        num_elements = preds_teacher.shape[1]
        num_elements_topk = int(np.ceil(num_elements / 20))  # top 5%
        weights_batch = torch.sum(torch.topk(preds_teacher,
        num_elements_topk).values, dim=1)
    return weights_batch

\end{python_l}

\section{Limitations}
\label{appendix:limitations}
In this section we will discuss some of the limitations of our proposed pipeline for pretraining on ImageNet-21K:

1) While our work did put a large emphasis on the efficiency of the proposed pretraining pipeline, for reasonable training times we still need an 8-GPUs machine (1 GPU training will be quite long, 2-3 weeks).

2) For creating an efficient pretraining scheme, and also to stay within our inner computing budget, we did not incorporate training tricks that significantly increase training times, although some of these tricks might give additional benefits and improve pretraining quality.

An example - techniques for dealing with extreme multi-tasking, such as GradNorm \cite{chen2018gradnorm} and PCGrad \cite{yu2020gradient}, that would probably improve the pretrain quality of multi-label training, but would significantly increase training times.

Another example of methods from the literature we have not tested - general "semantic" techniques that can be used for training neural networks (\cite{brust2019integrating,trammell2019contextual} for example). We found that most of these techniques are not feasible for large-scale efficient training. In addition, we believe that since our novel method, semantic softmax, is designed and tailored to the specific needs and characterizations of ImageNet-21K, it will significantly outperform general semantic methods.
\end{appendices}

3) When using private datasets which are larger than ImageNet-21K, such as JFT-300M \cite{sun2017revisiting}, the pretrain quality that can be achieved is probably still higher than the one we offer.

\end{document}